\pdfoutput=1

\documentclass[11pt]{article}
\usepackage{multirow}
\usepackage[table,xcdraw]{xcolor}
\usepackage{graphicx}
\usepackage{enumitem}
\usepackage{footnote}
\makesavenoteenv{tabular}

\usepackage{acl}

\usepackage{times}
\usepackage{latexsym}

\usepackage[T1]{fontenc}

\usepackage[utf8]{inputenc}

\usepackage{microtype}
\usepackage{booktabs}
\title{Perturbations and Subpopulations for Testing Robustness in Token-Based Argument Unit Recognition}

\author{Jonathan Kamp$^{1}$\hspace{1cm}Lisa Beinborn$^{1}$\hspace{1cm}Antske Fokkens$^{1,2}$ \\
    $^{1}$Computational Linguistics and Text Mining Lab, Vrije Universiteit Amsterdam \\
    $^{2}$Dept. of Mathematics and Computer Science, Eindhoven University of Technology \\
    \texttt{\{j.b.kamp,l.beinborn,antske.fokkens\}@vu.nl}}

\begin{document}
\maketitle
\begin{abstract}
Argument Unit Recognition and Classification aims at identifying argument units from text and classifying them as \textit{pro} or \textit{against}. One of the design choices that need to be made when developing systems for this task is what the unit of classification should be: segments of tokens or full sentences. Previous research suggests that fine-tuning language models on the token-level yields more robust results for classifying sentences compared to training on sentences directly. We reproduce the study that originally made this claim and further investigate what exactly token-based systems learned better compared to sentence-based ones. We develop systematic tests for analysing the behavioural differences between the token-based and the sentence-based system. Our results show that token-based models are generally more robust than sentence-based models both on manually perturbed examples and on specific subpopulations of the data.

\end{abstract}

\section{Introduction}

\begin{figure*}
\centering
\includegraphics[scale=0.30]{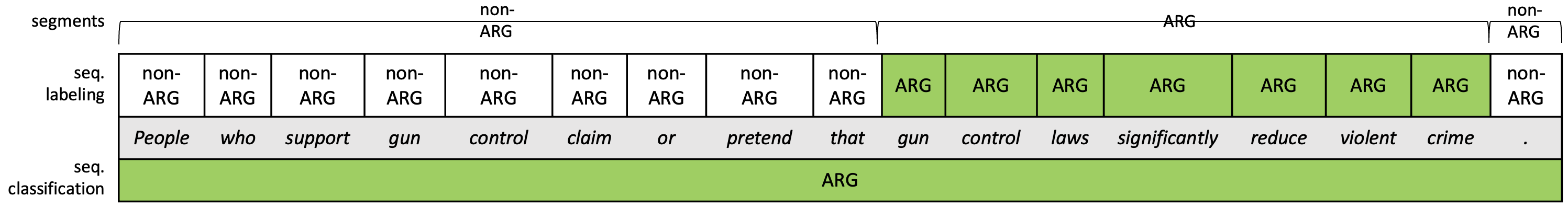}
\caption{An example sentence from the AURC-8 \citep{trautmann2020fine} topic \textit{gun control}. Each sentence in the dataset has one vector of token-wise gold labels (in- and output in a sequence \textit{labeling} approach, i.e.\ \textbf{token-based}) as well as one sentence-wise gold label (in- and output in a sequence \textit{classification} approach, i.e. \textbf{sentence-based}). ARG and \underline{non-}ARG gold \textit{segments} are sequences of tokens that carry the same label.}
\label{fig:figure1}
\end{figure*}

Identifying argumentation units is difficult, both for humans and machines. The challenge starts with the question of what it means for a segment to be argumentative towards a given topic in the first place \cite[e.g.]{trautmann2020fine,habernal2014argumentation}.
\citet{trautmann2020fine} propose a pragmatic approach for defining arguments and ask annotators to identify segments that can be placed in the $<$argument span$>$ slot of the following template: \textit{“$<$TOPIC$>$ should be supported/opposed, because $<$argument span$>$”}. They compare models that are trained to label tokens as being part of argumentative segments to models that classify full sentences as containing argumentative segments (ARG) or not (\underline{non-}ARG) (see Figure \ref{fig:figure1}), ultimately arguing that token-based training is preferable. Their experiments suggest that a token-based approach is more robust when sentence boundaries are unknown or not precisely given.

Argumentative segments provide reasons for taking a positive (pro) or negative (against) stance on a topic. These arguments are highly topic-specific, but the decent accuracy of cross-topic models indicates that there are also topic-independent cues. \citet{niven-kao-2019-probing} previously showed that transformer-based models learn to map specific cue words to a label and learn little about argumentation reasoning. It could be that this is the most we can expect in a cross-topic scenario. The question then remains where these cues are found: are they in the ARG segments themselves or are they also provided by the \underline{non-}ARG context? When comparing token-based and sentence-based models, we expect token-based models to be better at picking up cues that are specific to ARG segments themselves, whereas sentence-based models may be more susceptible to cues from the \underline{non-}ARG context, in particular, when these appear to announce an argumentation (e.g.\ \textit{because I think that...}). Reliance on (\underline{non-}ARG) cues is a particularly strong signal that general cues rather than reasoning are used.

In this paper, we dive further into this line of research. 
We rerun experiments with the best models of \citet{trautmann2020fine} to ensure a fair basis of comparison, reproducing most of the original results and coming close for the rest. We then design multiple robustness tests comparing the behavior of token- and sentence-based models in mixed-segment sentences, i.e.\ sentences that contain at least one ARG segment and one \underline{non-}ARG segment. We expect token-based models to be more robust because they are trained to distinguish between ARG and \underline{non-}ARG segments within sentence boundaries and thus have access to more precise information as to what makes up an ARG segment during training. This hypothesis is confirmed by our perturbation tests, which also show different behavior on subpopulations of the data. 
We thus show that a relatively small, curated dataset of adversarial examples can provide systematic insights into model behavior. Additional robustness tests with subpopulations of the data surprisingly do not yield clear differences between the two approaches.

\section{Background and Related Work}\label{Background}
In this section, we first present related work on argument unit recognition (\S\ref{RelatedWork}) and then dive further into the concept of robustness tests (\S\ref{WhatisRobustnessTesting}).

\subsection{Argument Unit Recognition}\label{RelatedWork}
Argumentation theory is about identifying how humans reach common ground and compromise, how societal information is exchanged, what the degree of subjectivity in viewpoints is and how polarised different stances can be. In the digital era, arguments from a wide range of sources are analysed. These sources range from debates on social media and (online) fora to technical documents used by professionals in the legal domain. Arguments roughly reflect the rationale behind a stance or decision, in relation to a certain topic or proposition. The field of computational argumentation attempts to model the argument patterns that are present in human language. \citet{lauscher2021scientia} distinguish between different tasks in argument modeling: $\sim$mining, $\sim$assessment, $\sim$reasoning, and $\sim$generation. Argument Unit Recognition and Classification is a task that can be positioned within argument \textit{mining}, as argumentative from non-argumentative expressions are first distinguished, and a stance is then attributed to each of the identified arguments. \citet{ajjour2017unit} show that the task of segmenting a text into argument units of different types remains particularly challenging in a cross-domain setting.

The first part of this research aims to reproduce the Argument Unit Recognition and Classification experiments by \citet{trautmann2020fine}, who train multiple transformer-based models on a novel argumentation dataset that is labeled at the token level: spans of tokens are then predicted as being pro, against or non-argumentative towards a given topic.

Argument mining has been thoroughly approached by (Bi-)LSTM modeling \citep{eger-etal-2017-neural}, SVMs and RNNs \citep{niculae-etal-2017-argument}. Apart from \citet{trautmann2020fine}, however, transformer-based architectures have been deployed more rarely. \citet{poudyal-etal-2020-echr} show how RoBERTa \citep{liu2019roberta} can successfully be applied on the legal ECHR dataset on a claim-premise task. \citet{ruiz2021transformer} test several flavors of BERT models \citep{devlin-etal-2019-bert} on the same task, but on a less domain-specific debate corpus. \citet{mayer2020transformer} compare different domain-generic and -specific transformer-based models in combination with CRF and GRU layers on medical texts. Similarly to \citet{trautmann2020fine}, they experiment with both sequence labeling and sequence classification, applying the former to a component detection task, and using the latter to classify relations between argument components. 

The next subsection provides background and related work on the second contribution of our paper: testing robustness.

\subsection{Robustness Testing}\label{WhatisRobustnessTesting}
\citet{goel-etal-2021-robustness} describe three ways of testing robustness: (1) testing on \textbf{subpopulations} of the test data the model is expected to perform poorly on; (2) \textbf{perturbing} the test data by creating adversarial examples \citep{zhang2020adversarial} that are expected to shed light on weaknesses of the model; (3) assessing model performance on pre-existing evaluation sets to establish \textbf{scalability} and \textbf{cross-domain} validity. We briefly discuss the role of each of these three in our work.

We design three \textbf{subpopulation} tests, two of which are based on similarity between training and test instances and one based on the ratio of argumentative tokens in a sentence. We also design three \textbf{perturbation} tests. The first design choice involves the level of granularity, which is usually on the word- or phrase-level. In our case, perturbation units are aligned with the granularity of the annotated spans, i.e.\ the ARG or \underline{non-}ARG segments remain intact but are combined in different ways. A second point of attention in creating perturbations is the risk of altering the grammar or semantics in an unintended way. Automatic metrics have been utilised to determine whether linguistic aspects are preserved after perturbation, such as the Jaccard similarity coefficient, grammar and syntax related measurements and edit-based measurements \citep{zhang2020adversarial}. These may be relatively fast to use, especially on a large scale, but might fall short in tasks where generating adversarial candidates goes beyond relatively simple, single word substitutions. We therefore opt for manual verification of our samples. \citet{trautmann2020fine} include the third type of robustness test already in that they apply \textbf{cross-topic} evaluation. Since we are mostly interested in the models' generic ability in identifying argumentative segments, we apply our robustness tests in the cross-topic setting only.

A handful of studies have applied robustness tests to transformer-based models on an argument mining task. \citet{schiller2021stance} apply paraphrases, spelling alterations and negation stress tests on a stance detection task. \citet{niven-kao-2019-probing} apply a negation stress test on a Argument Reasoning Comprehension Task, where negating a warrant (i.e.\ a type of argument) should result in predicting the inverted label. \citet{mayer2020generating} protract robustness testing into adversarial training: by inserting or replacing simple linguistic elements in the original data, such as nouns, scalar adverbs and punctuation, they use the perturbed examples for retraining the model, achieving higher performance. \citet{mayer2020generating} show the effectiveness of single, token-level perturbations, while aiming to control for \textit{same-meaning} preservation between the original and perturbed example pairs. Instead, we focus on the different argumentative load that different parts of a sentence carry to guide our perturbations, and ensure that the result of each perturbation is semantically sound (yet not unaltered). Finally, the in-domain versus cross-domain comparison is a more frequent type of testing, but it is often approached from a generalisability perspective (\textit{how well does the model perform on cross-domain data?}), which has a slightly different connotation from a robustness perspective (\textit{how well does the model defend itself from specific adversaries in cross-domain data?}). Our work can be seen as an extension to \citet{trautmann2020fine}, who find that token-based models are more robust against sentence segmentation errors than sentence-based models. Our robustness tests go beyond their work in that they show that token-based models are also more robust compared to sentence-based models in well-formed sentences with manipulated combinations of ARG and \underline{non-}ARG segments. Furthermore, the phenomena that we are testing robustness on are more likely to occur than scenarios in which sentence boundaries are not given.

\section{Reproduction Experiments}\label{ReproductionExperiments}
This section describes our reproduction study, including the dataset used (\S\ref{DatasetDescription}), the experimental setup (\S\ref{ExperimentalSetup}), the model evaluation metrics (\S\ref{ModelEvaluation}), and the requirements for a successful reproduction together with our results (\S\ref{ReproductionResultsAndConsiderations}).
\subsection{Dataset Description}\label{DatasetDescription}
The AURC-8 dataset developed by \citet{trautmann2020fine} is divided over eight topics: \textit{1. abortion 2. cloning 3. marijuana legalization 4. minimum wage 5. nuclear energy 6. death penalty 7. gun control 8. school uniforms}. In their manual labeling process, annotators were presented with candidate sentences in which arguments related to one given topic were possibly present. Argument spans were annotated according to the slot-filling template
\textit{“$<$TOPIC$>$ should be supported/opposed, because $<$argument span$>$”}. This results in spans annotated as $PRO$ (a supporting argument) or $CON$ (an opposing argument). Spans that remain unlabeled are assigned $NON$ (a non-argumentative segment). As an example, both underlined spans in the following sentence about \textit{death penalty} are labeled as $CON$ segments: `\textit{\underline{It does not deter crime} and \underline{it is extremely expensive to administer} .}' Instead, the first underlined span in the following sentence about \textit{gun control} is labeled as a $CON$ segment whereas the second span is labeled as $PRO$: `\textit{Yes , \underline{guns can be used for protection} but \underline{laws are meant to protect us} , too .}' In both example sentences, the spans of adjacent non-underlined tokens form the $NON$ segments. The dataset consists of 1,000 example sentences per topic. Of the 8,000 total sentences, 3,500 (43.75\%) are annotated as ARG and 4,500 (56.25\%) as \underline{non-}ARG. The portion of ARG sentences is divided over 658 examples (14.62\%) containing exclusively $PRO$ segments, 621 examples (13.80\%) containing exclusively $CON$ segments and 3,221 (71.58\%) containing any combination of $PRO$, $CON$ and $NON$ segments.\footnote{In this calculation, $NON$ segments that are solely formed by punctuation marks are ignored.}

The models are run on two different splits of the data: in-domain and cross-domain. In the in-domain setup, the first 70\% of the examples from each of the Topics 1-6 is assigned to training, the next 10\% to the development set, and the last 20\% to the test set. The cross-domain setup assigns all sentences from Topics 1-5 to training, Topic 6 to development, whereas Topic 7 and 8 form the test set.\footnote{Visit Appendix \ref{sec:appendix} for additional details on dataset versioning and pre-processing of the data.}

\subsection{Experimental Setup}\label{ExperimentalSetup}
We use two training approaches: token-based and sentence-based. 
\paragraph{Token-based} Models are trained on the sequence of token-wise gold labels, in a sequence-labeling fashion. The input to the model are tokenised sentences.
\paragraph{Sentence-based} Models are trained on a sentence-level gold label, in a sequence-classification fashion. The sentence-level gold label is a modification of the token-level gold labels. Let $t_L$ be the set of labels assigned to individual tokens in a sentence, and $f_{PRO}$ and $f_{CON}$ the number of tokens in the sentence that are labeled as $PRO$ and $CON$, respectively. Then, the sentence label $s_L$ is obtained as follows: 

\ \\
\begin{tabular}{ll}
$t_L = \{NON\},$ & $s_L:=NON$ \\ 
$t_L = \{NON,PRO\},$ & $s_L:=PRO$ \\
$t_L = \{NON,CON\},$ & $s_L:=CON$ \\
$t_L \supseteq \{PRO,CON\}$: \\
\qquad if $f_{PRO} > f_{CON},$ & $s_L:=PRO$ \\
\qquad if $f_{CON} > f_{PRO},$ & $s_L:=CON$ \\
\qquad if $f_{PRO} = f_{CON},$ & $s_L:=random$\footnote{A random choice from $\{PRO,CON\}$ is made.}
\end{tabular} \\

The input instance fed to the sentence-based model is the same tokenised sentence used as input in the token-based model. Instead of feeding along a sequence of token-wise labels, we feed its unique $s_L$. The output is a predicted $s_L$.

We re-train the models based on the architecture that performed best in the original paper: BERT{\tiny LARGE} \citep{devlin-etal-2019-bert}. We also train a token-based model with a CRF layer.\footnote{We were not able to re-implement the CRF layer for the sentence-based approach and could therefore not include this.} In the original results, the CRF layer improved segmentation. The model without CRF more often broke segments up into multiple single-word segments.

For each domain split, for each model setup we carry out series of 5 training runs with a different random seed for each run. We report mean F1-scores and standard deviation for each series of runs.
Hyperparameter settings are reported in Appendix \ref{sec:appendix}.

\subsection{Model Evaluation}\label{ModelEvaluation}
The models are evaluated on two metrics: token-F1 and sentence-F1.\footnote{\citet{trautmann2020fine} also include a third metric: segment-F1. Given that the description of their implementation remains underspecified and since the metric is not strictly relevant to our work, we report the original segment-F1 results in Appendix \ref{sec:appendix} along with our own segment-F1 implementation.} Token-F1 is calculated as the average over the per-class F1-scores for all tokens in the evaluation set. Sentence-F1 is the average over per-class F1-scores for all sentences in the evaluation set. Whereas token-F1 is straightforward for the token-based setup, and sentence-F1 is for the sentence-based setup, one extra step is needed to retrieve the sentence labels from the token-based predictions, and token labels from the sentence-based predictions. When using the token-based model, we obtain the sentence labels from the assigned tokens using the same approach as described in \S\ref{ExperimentalSetup}. After applying the sentence-based model, we obtain token labels by assigning the predicted sentence label to all tokens of the sentence.

\begin{table*}
\centering
\begin{tabular}{cllcccc}
\toprule
\multicolumn{3}{l}{} &
  \multicolumn{2}{c}{\textbf{\small token-F1}} &
  \multicolumn{2}{c}{\textbf{\small sentence-F1}} \\ \midrule
\multicolumn{2}{c}{\textit{\small setting}} &
  \multicolumn{1}{c}{\textit{\small model}} &
  \multicolumn{1}{c}{\textit{\begin{tabular}[c]{@{}c@{}}{\small token}\\{\small -based setup}\end{tabular}}} &
  \multicolumn{1}{c}{\textit{\begin{tabular}[c]{@{}c@{}}{\small sentence}\\{\small -based setup}\end{tabular}}} &
  \multicolumn{1}{c}{\textit{\begin{tabular}[c]{@{}c@{}}{\small token}\\{\small -based setup}\end{tabular}}} &
  \multicolumn{1}{c}{\textit{\begin{tabular}[c]{@{}c@{}}{\small sentence}\\{\small -based setup}\end{tabular}}} \\ \midrule
\multicolumn{1}{c}{} &
  \multicolumn{1}{l}{} &
  {\small BERT}{\tiny LARGE} &
  \multicolumn{1}{c}{\small.683} &
  {\small.627} &
  \multicolumn{1}{c}{\small.709} &
  {\small.715} \\ 
\multicolumn{1}{c}{} &
  \multicolumn{1}{l}{\multirow{-2}{*}{\small orig}} &
  {\small BERT}{\tiny LARGE}{\small+CRF} &
  \multicolumn{1}{c}{\small.696} &
  {\small.622} &
  \multicolumn{1}{c}{\small.711} &
  {\small.725} \\  
\multicolumn{1}{c}{} &
  \multicolumn{1}{l}{\cellcolor[HTML]{FFFFC7}} &
  \cellcolor[HTML]{FFFFC7}{\small BERT}{\tiny LARGE} &
  \multicolumn{1}{c}{\cellcolor[HTML]{FFFFC7}\begin{tabular}[c]{@{}l@{}}{\small.698} {\tiny(.003)}\end{tabular}} &
  \cellcolor[HTML]{FFFFC7}\begin{tabular}[c]{@{}l@{}}\textbf{\small.614} {\tiny(.008)}\end{tabular} &
  \multicolumn{1}{c}{\cellcolor[HTML]{FFFFC7}\begin{tabular}[c]{@{}l@{}}\textbf{\small.708} {\tiny(.004)}\end{tabular}} &
  \cellcolor[HTML]{FFFFC7}\begin{tabular}[c]{@{}l@{}}\textbf{\small.713} {\tiny(.012)}\end{tabular} \\ 
\multicolumn{1}{c}{\multirow{-4}{*}{\small in-domain}} &
  \multicolumn{1}{l}{\multirow{-2}{*}{\cellcolor[HTML]{FFFFC7}{\small repr}}} &
  \cellcolor[HTML]{FFFFC7}{\small BERT}{\tiny LARGE}{\small+CRF} &
  \multicolumn{1}{c}{\cellcolor[HTML]{FFFFC7}\begin{tabular}[c]{@{}l@{}}\textbf{\small.696} {\tiny(.003)}\end{tabular}} &
  \cellcolor[HTML]{FFFFC7}- &
  \multicolumn{1}{c}{\cellcolor[HTML]{FFFFC7}\begin{tabular}[c]{@{}l@{}}\textbf{\small.711} {\tiny(.006)}\end{tabular}} &
  \cellcolor[HTML]{FFFFC7}- \\ 
\multicolumn{1}{c}{} &
  \multicolumn{1}{l}{} &
  {\small BERT}{\tiny LARGE} &
  \multicolumn{1}{c}{\small.596} &
  {\small.544} &
  \multicolumn{1}{c}{\small.598} &
  {\small.602} \\  
\multicolumn{1}{c}{} &
  \multicolumn{1}{c}{\multirow{-2}{*}{\small orig}} &
  {\small BERT}{\tiny LARGE}{\small+CRF} &
  \multicolumn{1}{c}{\small.620} &
  {\small.519} &
  \multicolumn{1}{c}{\small.610} &
  {\small.573} \\ 
\multicolumn{1}{c}{} &
  \multicolumn{1}{c}{\cellcolor[HTML]{FFFFC7}} &
  \cellcolor[HTML]{FFFFC7}{\small BERT}{\tiny LARGE} &
  \multicolumn{1}{c}{\cellcolor[HTML]{FFFFC7}\begin{tabular}[c]{@{}l@{}}\textbf{\small.587} {\tiny(.008)}\end{tabular}} &
  \cellcolor[HTML]{FFFFC7}\begin{tabular}[c]{@{}l@{}}\textbf{\small.529} {\tiny(.011)}\end{tabular} &
  \multicolumn{1}{c}{\cellcolor[HTML]{FFFFC7}\begin{tabular}[c]{@{}l@{}}\textbf{\small.604} {\tiny(.009)}\end{tabular}} &
  \cellcolor[HTML]{FFFFC7}\begin{tabular}[c]{@{}l@{}}{\small.566} {\tiny(.017)}\end{tabular} \\ 
\multicolumn{1}{c}{\multirow{-4}{*}{\small cross-domain}} &
  \multicolumn{1}{c}{\multirow{-2}{*}{\cellcolor[HTML]{FFFFC7}{\small repr}}} &
  \cellcolor[HTML]{FFFFC7}{\small BERT}{\tiny LARGE}{\small+CRF} &
  \multicolumn{1}{c}{\cellcolor[HTML]{FFFFC7}\begin{tabular}[c]{@{}l@{}}{\small.578} {\tiny(.008)}\end{tabular}} &
  \cellcolor[HTML]{FFFFC7}- &
  \multicolumn{1}{c}{\cellcolor[HTML]{FFFFC7}\begin{tabular}[c]{@{}l@{}}\textbf{\small.609} {\tiny(.007)}\end{tabular}} &
  \cellcolor[HTML]{FFFFC7}- \\ \bottomrule
\end{tabular}
\caption{Original results (white background) compared to reproduction results (non-white background) on the test set. Models are divided over an in- and cross-domain setting. Reproduction results show the mean scores from 5 runs, along with the standard deviation (within parentheses). The reproduction scores where the original score falls within two standard deviations from the mean are given in bold.}
\label{fig:reproductionresults}
\end{table*}

\subsection{Reproduction Results and Considerations}\label{ReproductionResultsAndConsiderations}

We consider the mean F1-scores over three runs from \citet{trautmann2020fine} as the benchmark for the reproduction comparisons. We follow \citet{moore-rayson-2018-bringing} and provide F1-distributions reporting the mean and standard deviation from our experiments. It remains a methodological challenge to determine a threshold within which a score can be defined as successfully reproduced. We follow \citet{reuver-etal-2021-stance} and consider the reproduction successful if  given a distribution of reproduced F1-scores $D$, the original mean F1-score falls within two standard deviations from the mean of $D$. We provide all individual decisions on the test set for a more accurate comparison, since F1-scores can still stem from different behavior on sub-populations of the data. 

At a first glance, the differences between the original and replicated results are relatively small for both token-based and sentence-based models. One pattern from the original paper is not reproduced, namely, the positive effect on performance by the CRF layer on the token-based model. 
In the light of the threshold of two standard deviations, we observe in Table~\ref{fig:reproductionresults} that reproductions are partially successful. On the test set, 5 out of 6 F1-scores are reproduced in the in-domain setting, and 4 out of 6 in the cross-domain setting. Success rate of reproduction does not seem to depend on training setup either: 6 out of 8 for token-based versus 3 out of 4 for sentence-based. Scores that are not reproduced come close as they fall within half a decimal from the original.\footnote{See Table \ref{fig:reproductionresultscomplete} in Appendix \ref{sec:appendix} for a complete overview of the reproduction results. It can be observed that none of the segment-F1 metrics are reproduced, probably caused by a slightly different implementation of how these scores are calculated.}


\section{Robustness Testing}\label{RobustnessTesting}
We test robustness in a cross-domain setting. By isolating this problem from topic-dependent content biases, the models are expected to focus more on indicators that are representative of a generic notion of argumentation. While a token-based model is explicitly instructed that there are fine-grained argumentative differences within a sentence, a sentence-based model is not. Therefore, we expect the sentence-based models to have more difficulty in predicting the cues that are argumentative on a micro-level (i.e.\ tokens, segments), which translates to difficulties at the macro-level (i.e.\ the sentence). Our robustness tests precisely operate at a micro-level: adding, replacing or removing segments should impact the sentence-based model more negatively than the token-based model. 

We apply robustness tests to the two cross-domain token-based models (BERT{\tiny LARGE}, BERT{\tiny LARGE}{\small+CRF}) and the sentence-based model (BERT{\tiny LARGE}). We investigate robustness for the task as a binary prediction problem (Argument Unit Recognition) and remove the stance component: ARG entails both labels $PRO$ and $CON$, and \underline{non-}ARG corresponds to $NON$. As anticipated in \S\ref{WhatisRobustnessTesting}, we categorise the robustness tests according to two classes: \textit{perturbations} on the test set (\S\ref{perturbations}) and \textit{subpopulations} of the test set (\S\ref{subpopulations}).

\subsection{Perturbations on the Test Set}\label{perturbations}
We craft a $before$-dataset and $after$-dataset in the following way. First, artificial candidate test sets are generated through deletion, recombination or label-based pre-selection of segments. The segments are sampled from the original test set. Second, we manually label or complete the candidate examples. We create three types of tests \textbf{T1}, \textbf{T2} and \textbf{T3}.
 We report on the impact of the perturbation through $\Delta acc$, i.e.\ the difference between the accuracy \textit{before} and the accuracy \textit{after} the perturbation has been applied. Hence, each example in either the \textit{before}- or \textit{after}-dataset has one gold label (at the sentence-level) on which the models are evaluated. 

\paragraph{T1 - Announcing Segments} Observations in the original test set show that \underline{non-}ARG segments can broadly be divided in segments that announce (ANN) an immediately subsequent ARG segment, and segments that do not (non-ANN). For instance, ANN segments are phrases the include literal argument indicators such as \textit{evidence, claim, argument, reason} followed by a copula, and phrases that include reporting verbs. Examples:
\begin{itemize}
    \item []\textbf{ANN}\\ \textit{…a major argument against this topic is…}\\ \textit{…he thinks that…}
    \item []\textbf{non-ANN}\\ \textit{…this document was written in 2022 and…}\\ \textit{…but…}
\end{itemize}
ANN segments are an example of information that is known to be \underline{non-}ARG by the token-based model, but not by the sentence-based model where it falls under a coarse-grained, sentence-level ARG label.
Since ANN segments mostly co-occur with ARG segments, the sentence-based model is likely to mix them up. The token-based model may also use an ANN segment as signal that an ARG is following, but has better chances of using information from the following segment itself to identify when this is not the case. We test this by creating counter-examples that concatenate ANN segments to a subsequent \textit{non-}ARG segment. This results in \underline{non-}ARG sentence-level labels, for instance, \textit{`Pro-abortion politicians think that...' + `...the debate has become very delicate.'}. If our theory is correct, the token-based model would generally be able to classify the two segments separately as \underline{non-}ARG, resulting in a \underline{non-}ARG label for the sentence, whereas the sentence-model is more prone to label the sentence as ARG based on the ANN segment.

\ \\
\begin{center}
\begin{tabular}{lrr}
\toprule
                & \textbf{\small concatenation}                                                  & \textbf{\small sentence gold} \\ \midrule
\textit{\small before} & \begin{tabular}[c]{@{}r@{}}{\small ANN \underline{non-}ARG seg.}\\ {\small + ARG seg.}\end{tabular}     & {\small ARG}                    \\ \midrule
\textit{\small after}  & \begin{tabular}[c]{@{}r@{}}{\small ANN \underline{non-}ARG seg.}\\ {\small + \underline{non-}ARG seg.}\end{tabular} & {\small \underline{non-}ARG}                \\
\bottomrule
\end{tabular}
\end{center}

We first extracted candidate $<a,b>$ pairs, where $a$ is an ARG segment, $b$ is a \underline{non-}ARG segment and $a$ is immediately followed by $b$ in the same sentence from the original AURC-8 dataset. Pairs that do not form a full sentence are manually discarded. Subsequently, we manually labeled the \underline{non-}ARG segments as (non-)ANN, until reaching 100 ANN annotations for the \textit{gun control} topic and 100 for \textit{school uniforms}. Each ANN segment (e.g.\ \textit{`Pro-abortion politicians think that...'}) is then manually completed with a novel \underline{non-}ARG segment (e.g.\ \textit{`...the debate has become very delicate.'}) to form a full \underline{non-}ARG sentence and is added to the \textit{after}-dataset. The respective $<a,b>$ pairs are added to the \textit{before}-dataset. 

\paragraph{T2 - Concatenate Non-Argumentative Sentence} Here we test robustness by concatenating an ARG segment with a pure \underline{non-}ARG sentence. In between the two segments, the connector \textit{`and besides,'} is used to create a well-formed sentence. This results in constructions where the ARG segment ends up in a context of a relatively high number of \underline{non-}ARG tokens. Such a concatenation would result in e.g.: \textit{`Uniforms force conformity'+ `and besides,' + `it's a great service for parents as I was able to pick up lots of good stuff for little money.'} The token-based model is expected to classify the two segments as ARG and \underline{non-}ARG respectively, resulting in a sentence-wise ARG label prediction. The sentence-based model might be more biased by the high ratio of \underline{non-}ARG tokens that are present in the sentence, potentially resulting in a sentence-wise \underline{non-}ARG prediction.

\ \\
\begin{center}
\begin{tabular}{lrr}
\toprule
                & \textbf{\small concatenation}                                             & \textbf{\small sentence gold} \\ \midrule
\textit{\small before} & {\small ARG seg.}                                                           & {\small ARG}                    \\ \midrule
\textit{\small after}  & \begin{tabular}[c]{@{}r@{}}{\small ARG seg.}\\ {\small + connector} \\ {\small + \underline{non-}ARG sent.}\end{tabular} & {\small ARG}                    \\ \bottomrule
\end{tabular}
\end{center}
\ \\

 We populate a candidate dataset with concatenations of an ARG segment, the connector and a pure \underline{non-}ARG sentence, in that order. The components in each concatenation (except for the connector, which is constant) are on the same topic and are sampled from the original test set. The ARG segment should be a full stand-alone sentence. From this candidate dataset, we then select 50 examples for \textit{gun control} and 50 for \textit{school uniforms} that are sound, stand-alone sentences to be added to the \textit{after}-dataset. The \textit{before}-dataset consists of the respective ARG segments.

\paragraph{T3 - Remove Non-Argumentative Segment} In this test, we remove the \underline{non-}ARG context around the remaining ARG segment creating uncontextualised argument units. We expect this perturbation to have less impact on the token-based model, as its decision is potentially less informed by the missing \underline{non-}ARG segments. 

\ \\
\begin{center}
\begin{tabular}{lrr}
\toprule
                & \textbf{\small concatenation}                                                                               & \textbf{\small sentence gold} \\ \midrule
\textit{\small before} & \begin{tabular}[c]{@{}r@{}}{\small ARG seg.} \\ {\small + \underline{non-}ARG seg.} \\ {\small/}\\ {\small \underline{non-}ARG seg.} \\ {\small + ARG seg.}\end{tabular} & {\small ARG}                    \\ \midrule
\textit{\small after}  & {\small ARG seg.}                                                                                             & {\small ARG}                    \\ \bottomrule
\end{tabular}
\end{center}
\ \\

We extract pairs consisting of an ARG and a \underline{non-}ARG segment from the original corpus. Both elements of each pair stem from the same source sentence and are originally adjacent. We manually check them to ensure that both the pair and the ARG segment alone form well-formed sentences. We select a total of 200 examples with an approximate 50\%-50\% split of examples where \underline{non-}ARG precedes or follows the ARG segment, as well as an approximate 50\%-50\% split between the two topics. The pairs form the \textit{before}-dataset whereas the ARG segments alone form the \textit{after}-dataset.

\subsection{Subpopulations of the Test Set}\label{subpopulations}
A subpopulation is a group of test instances that is selected based upon a criterion that is expected to influence the performance of the model. We take it a step further: we consider each instance in the test set a subpopulation on its own and assign it a value from a continuous variable in the data. In our case, the continuous variable is a semantic similarity score between train and test data, and the ratio of noisy (non-argumentative) tokens per sentence, two aspects that generally impact language classification tasks. The point-biserial correlation coefficient $r_{pb}$ is then calculated between this continuous variable and the dichotomous prediction correctness. Thus, $r_{pb}$ is expected to be lower for a model when the continuous variable forms less of a bias on its decisions compared to its effect on another model.

\paragraph{T4 - Similarity Train-Test Same Labels} The outcome of this test provides an indication of the impact of semantic similarity between training and test data on the decision of the model. For each of the mixed-segment sentences in the test set, a pair-wise semantic similarity coefficient is calculated in relation to each of the sentences in the training set. If the maximum semantic similarity coefficient for one test sentence corresponds to a training instance with the same label (ARG), the test sentence is stored in the \textit{T4-set} along with its coefficient. The correlation between prediction correctness of mixed-segment sentences from the \textit{T4-set} and their respective maximum similarity coefficients is then computed. We expect the sentence-based model to be more affected by it than the token-based model, given that semantic similarity at the macro-level of the sentence may be a more prominent indicator for the former model. A token-based setup, on the other hand, should be able to classify segments within the sentence as it can rely on explicit ARG vs \underline{non-}ARG information. This translates to a correlation coefficient that is expected to be higher for the sentence-based model than for the token-based model.

The semantic vector representation of a sentence is given by its averaged token vectors.\footnote{\url{spacy.io/models/en} $\rightarrow$ \texttt{en\_core\_web\_lg v3.3.0}.} Sentence similarity corresponds to the cosine similarity between the two semantic vector representations of a sentence pair. 

\paragraph{T5 - Similarity Train-Test Opposite Labels} For T5, the maximum similarity coefficient is calculated in relation to the instances in the training set that have an opposite label (\underline{non-}ARG) to the mixed-segment sentences. Similarly to the \textit{T4-set}, a \textit{T5-set} is created accordingly. This aspect is expected to have more impact on the sentence-based model than the token-based model, hence yielding a weaker correlation for the latter.    
    
\paragraph{T6 - Argumentative Token Ratio} Through T6, prediction correctness is correlated with the argumentative token ratio in mixed-segment sentences from the test set. This ratio is calculated as the number of ARG tokens over the number of all tokens. In line with the expectations in T4 and T5, the token-based model should be less affected by this sentence-level aspect, resulting in a weaker correlation coefficient compared to the sentence-based model.
\begin{table*}[th!]
\begin{tabular}{llllllllll}
\toprule
\multicolumn{1}{c}{} & 
\multicolumn{3}{c}{\textbf{{\small T1}}} & 
\multicolumn{3}{c}{\textbf{{\small T2}}} & 
\multicolumn{3}{c}{\textbf{{\small T3}}} \\ \midrule 
\multicolumn{1}{l}{\textit{\small model}} & 
\multicolumn{1}{c}{\textit{\small before}} & 
\multicolumn{1}{c}{\textit{\small after}} & 
\multicolumn{1}{c}{\textit{{\small $\Delta acc$}}} & 
\multicolumn{1}{c}{\textit{\small before}} & 
\multicolumn{1}{c}{\textit{\small after}} & 
\multicolumn{1}{c}{\textit{{\small $\Delta acc$}}} & 
\multicolumn{1}{c}{\textit{\small before}} & 
\multicolumn{1}{c}{\textit{\small after}} & 
\multicolumn{1}{c}{\textit{{\small $\Delta acc$}}} \\ \midrule
\begin{tabular}[c]{@{}l@{}}{\small token-based}\\{\small BERT}{\tiny LARGE}\end{tabular} & 
 \multicolumn{1}{c}{\begin{tabular}[c]{@{}l@{}}{\small.875} {\tiny(.018)}\end{tabular}} &  
 \multicolumn{1}{c}{\begin{tabular}[c]{@{}l@{}}{\small.832} {\tiny(.033)}\end{tabular}} & 
 \multicolumn{1}{c}{\begin{tabular}[c]{@{}l@{}}\small{\textbf{-.043}}\\ \small{\textbf{-4.8{\scriptsize\%}}}\end{tabular}} &
 \multicolumn{1}{c}{\begin{tabular}[c]{@{}l@{}}{\small.760} {\tiny(.035)}\end{tabular}} &  
 \multicolumn{1}{c}{\begin{tabular}[c]{@{}l@{}}{\small.830} {\tiny(.037)}\end{tabular}} & 
 \multicolumn{1}{c}{\begin{tabular}[c]{@{}l@{}}\small{\textbf{+.070}}\\ \small{\textbf{+9.2{\scriptsize\%}}}\end{tabular}} &
 \multicolumn{1}{c}{\begin{tabular}[c]{@{}l@{}}{\small.760} {\tiny(.022)}\end{tabular}} & 
 \multicolumn{1}{c}{\begin{tabular}[c]{@{}l@{}}{\small.683} {\tiny(.037)}\end{tabular}} & 
 \multicolumn{1}{c}{\begin{tabular}[c]{@{}l@{}}\small{\textbf{-.077}}\\ \small{\textbf{-10.1{\scriptsize\%}}}\end{tabular}} \\ \midrule
\begin{tabular}[c]{@{}l@{}}{\small token-based}\\{\small BERT}{\tiny LARGE}{\small+CRF}\end{tabular} & 
\multicolumn{1}{c}{\begin{tabular}[c]{@{}l@{}}{\small.878} {\tiny(.021)}\end{tabular}} & 
\multicolumn{1}{c}{\begin{tabular}[c]{@{}l@{}}{\small.856} {\tiny(.024)}\end{tabular}} & 
\multicolumn{1}{c}{\begin{tabular}[c]{@{}l@{}}{\textbf{\small-.022}}\\ {\textbf{\small-2.5{\tiny\%}}}\end{tabular}} &
\multicolumn{1}{c}{\begin{tabular}[c]{@{}l@{}}{\small.790} {\tiny(.015)}\end{tabular}} & 
\multicolumn{1}{c}{\begin{tabular}[c]{@{}l@{}}{\small.760} {\tiny(.028)}\end{tabular}} & 
\multicolumn{1}{c}{\begin{tabular}[c]{@{}l@{}}{\textbf{\small-.030}}\\ {\textbf{\small-3.8{\scriptsize\%}}}\end{tabular}} &
\multicolumn{1}{c}{\begin{tabular}[c]{@{}l@{}}{\small.740} {\tiny(.036)}\end{tabular}} & 
\multicolumn{1}{c}{\begin{tabular}[c]{@{}l@{}}{\small.672} {\tiny(.023)}\end{tabular}} & 
\multicolumn{1}{c}{\begin{tabular}[c]{@{}l@{}}{\textbf{\small-.068}}\\ {\textbf{\small-9.2{\scriptsize\%}}}\end{tabular}} \\ \midrule
\begin{tabular}[c]{@{}l@{}}{\small sentence-based}\\{\small BERT}{\tiny LARGE}\end{tabular} & 
\multicolumn{1}{c}{\begin{tabular}[c]{@{}l@{}}{\small.835} {\tiny(.034)}\end{tabular}} & 
\multicolumn{1}{c}{\begin{tabular}[c]{@{}l@{}}{\small.808} {\tiny(.053)}\end{tabular}} & 
\multicolumn{1}{c}{\begin{tabular}[c]{@{}l@{}}\small{\textbf{-.027}}\\ \small{\textbf{-3.2{\scriptsize\%}}}\end{tabular}} &
\multicolumn{1}{c}{\begin{tabular}[c]{@{}l@{}}{\small.638} {\tiny(.059)}\end{tabular}} & 
\multicolumn{1}{c}{\begin{tabular}[c]{@{}l@{}}{\small.640} {\tiny(.064)}\end{tabular}} &
\multicolumn{1}{c}{\begin{tabular}[c]{@{}l@{}}\small{\textbf{+.002}}\\ \small{\textbf{+0.3{\scriptsize\%}}}\end{tabular}} &
\multicolumn{1}{c}{\begin{tabular}[c]{@{}l@{}}{\small.652} {\tiny(.047)}\end{tabular}} & 
\multicolumn{1}{c}{\begin{tabular}[c]{@{}l@{}}{\small.576} {\tiny(.034)}\end{tabular}} & 
\multicolumn{1}{c}{\begin{tabular}[c]{@{}l@{}}\small{\textbf{-.076}}\\ \small{\textbf{-11.7{\scriptsize\%}}}\end{tabular}} \\ \bottomrule
\end{tabular}
\caption{Impact perturbations on cross-domain token-based and sentence-based models. Mean accuracy and standard deviation (within parentheses) over 5 runs is reported for each model. Accuracy is calculated on the test set before applying the perturbation (\textit{before}) and after applying the perturbation (\textit{after}). $\Delta acc$ represents the absolute and relative (\%) difference between \textit{before} and \textit{after}.}
\label{fig:impactperturbations}
\end{table*}
\begin{table*}[h]
\centering
\begin{tabular}{llllllllll}
\toprule
\multicolumn{1}{c}{} & 
\multicolumn{3}{c}{\textbf{{\small T4}}} & 
\multicolumn{3}{c}{\textbf{{\small T5}}} & 
\multicolumn{3}{c}{\textbf{{\small T6}}} \\ \midrule
\multicolumn{1}{l}{\textit{\small model}} &
\multicolumn{3}{c}{\textit{{\small $r_{pb}$}}} & 
\multicolumn{3}{c}{\textit{{\small $r_{pb}$}}} & 
\multicolumn{3}{c}{\textit{{\small $r_{pb}$}}} \\ \midrule
\begin{tabular}[c]{@{}l@{}}{\small token-based}\\{\small BERT}{\tiny LARGE}\end{tabular} & \multicolumn{3}{c}{\begin{tabular}[c]{@{}l@{}}{\small-.068} {\tiny (.031)}\end{tabular}} & \multicolumn{3}{c}{\begin{tabular}[c]{@{}l@{}}{\small.027} {\tiny (.009)}\end{tabular}} & \multicolumn{3}{c}{\begin{tabular}[c]{@{}l@{}}{\small.028} {\tiny (.023)}\end{tabular}} \\ \midrule
\begin{tabular}[c]{@{}l@{}}{\small token-based}\\{\small BERT}{\tiny LARGE}{\small+CRF}\end{tabular} & \multicolumn{3}{c}{\begin{tabular}[c]{@{}l@{}}{\small-.031} {\tiny (.016)}\end{tabular}} & \multicolumn{3}{c}{\begin{tabular}[c]{@{}l@{}}{\small.013} {\tiny (.029)}\end{tabular}} & \multicolumn{3}{c}{\begin{tabular}[c]{@{}l@{}}{\small.046} {\tiny (.010)}\end{tabular}} \\ \midrule
\begin{tabular}[c]{@{}l@{}}{\small sentence-based}\\{\small BERT}{\tiny LARGE}\end{tabular} & \multicolumn{3}{c}{\begin{tabular}[c]{@{}l@{}}{\small-.014} {\tiny (.034)}\end{tabular}} & \multicolumn{3}{c}{\begin{tabular}[c]{@{}l@{}}{\small-.037} {\tiny (.044)}\end{tabular}} & \multicolumn{3}{c}{\begin{tabular}[c]{@{}l@{}}{\small.042} {\tiny (.037)}\end{tabular}} \\ \bottomrule
\end{tabular}
\caption{Impact subpopulations on cross-domain token-based and sentence-based models. \textit{r{\small pb}} indicates the point-biserial correlation coefficient between prediction correctness and a given aspect of the sentence. The range of \textit{r{\small pb}} is $[-1,1]$, where the two extremes indicate a perfect negative and positive correlation, respectively. The coefficients in the table are the means from 5 runs per model, along with the standard deviations (within parentheses).}
\label{fig:impactsubpopulations_v2}
\end{table*}

\subsection{Results Robustness Tests}
The perturbation results of T1, T2, T3 are collected in Table \ref{fig:impactperturbations}, where $\Delta acc$ quantifies the impact of each perturbation. Specifically, $\Delta acc$ represents the difference in accuracy by the models on sentences before and after the perturbation has been applied. It can be observed that an overall negative $\Delta acc$ pattern is present across the grid, which is expected behavior. The maximum absolute negative impact is $\Delta acc = -.077$, achieved through T3 on the token-based model without CRF layer. From a relative point of view, the sentence-based model is impacted most with -11.7\% on T3.

As an answer to our initial expectations, the token-based model with CRF layer is more robust to perturbations than the sentence-based model on two out of three tests: T1 (Announcing Segments) and T3 (Remove Non-Argumentative Segment). This is quantified in terms of both absolute $\Delta acc$ (-.022 on T1, -.068 on T3) and relative $\Delta acc$ (-2.5\% on T1, -9.2\% on T3). In comparison, the token-based model without CRF is impacted more heavily than the sentence-based model in absolute terms (-.043 versus -.027 on T1; -.077 versus -.076 on T3), but is more robust in relative terms on T3 (-10.1\% versus -11.7\%). Although the CRF layer has not proven to clearly increase the token-based model performance (not observable in Table \ref{fig:reproductionresults}, nor in Table \ref{fig:impactperturbations}), it appears to improve the robustness of the model.

Interestingly, the token-based model without CRF layer is the only one to considerably improve performance on the T2 \textit{after}-dataset. This behavior is unexpected since all \textit{after}-sets were meant to \textit{trick} the models rather than to help them. A possible explanation might be that the connector `\textit{and besides,}' is often included in the annotated ARG spans in AURC-8 training instances. This could represent a general downside of token-based models: picking up a small cue in the sentence as ARG, therefore predicting the sentence-wise label as ARG.

The results of subpopulation tests T4, T5 and T6 are given in Table~\ref{fig:impactsubpopulations_v2}. We hypothesised that continuous aspects in the data (such as semantic similarity between full sentences in training and test or the argument token ratio of a sentence) would correlate more strongly with predictions by the sentence-based model compared to the token-based model. This hypothesis could not be confirmed. Apart from being close to 0, which indicates no correlation, the \textit{r{\small pb}} coefficients for the token-based model are also close to the coefficients for the sentence-based model on the same tests T4-6, which indicates no difference in bias between the models. The perturbation results (T1-3), however, provide an indication that there are differences between subpopulations. Specifically, both token-based models achieve a higher accuracy on each single \textit{before-} and \textit{after-}dataset, which are specific subpopulations of the data. This clear difference in performance can be explained by the fact that these tests do not cover pure, non-argumentative sentences on which the sentence-based model might be stronger (as can be inferred from the comparable sentence-F1 scores between the two types of models in Table~\ref{fig:reproductionresults}). 
We therefore believe more research on subpopulations is needed. In particular, we may investigate alternative implementations of the continuous variables, such as using Sentence-bert \cite{reimers2019sentence} for representing the semantics of individual instances or also looking at the number of semantically similar examples in the training data.

\section{Conclusion}\label{Conclusion}

In this study, we partially reproduced the results of \citet{trautmann2020fine} and introduced new robustness tests that showed how token-based models are generally more robust than models trained at a sentence level on an Argument Unit Recognition task. We applied two type of tests: perturbations and subpopulations. With regards to the perturbations, we found that 1) removing the \underline{non-}ARG segment from a mixed-segment sentence, and 2) replacing the ARG segment with a \underline{non-}ARG segment in \textit{announcing} phrases such as `\textit{Their main argument is $<$ARG$>$}' or `\textit{Most politicians against gun legislation think that $<$ARG$>$}' negatively impact a sentence-based model more than a token-based model. We did not find a difference in bias among the two types of models with regards to semantic similarity between training and evaluation data, and high argumentative token ratios at the sentence level. Instead, we showed that the development of perturbation test sets itself can shed light on specific subpopulations of the data: our token-based models performed better on both mixed-segment sentences and single argumentative segments. 

By approaching the task from a challenging, cross-domain perspective, we isolated the problem from model reliance on topic-dependent content. Our analyses reveal that it is difficult to define a common denominator for the notion of argumentativeness across topics. They highlighted the importance of the type of knowledge we expect to be learned by a computational model of argumentation. Structural choices in the annotation setup can lead to systematic gaps in the dataset that allow the model to take superficial shortcuts \citep{gardner2020evaluating}. Robustness tests are a means to detect such gaps and, as a side effect, help in unraveling conceptual vagueness.  

\section*{Acknowledgements}
Jonathan Kamp's and Antske Fokkens' research was funded by the Dutch National Science Organisation (NWO) through the project InDeep: Interpreting Deep Learning Models for Text and Sound (NWA.1292.19.399). Lisa Beinborn’s research was funded by NWO through the project CLARIAH-PLUS (CP-W6-19-005).  
We would like to thank Dietrich Trautmann and his co-authors for making their clear code and dataset publicly available, and for the email correspondence. We also thank the anonymous reviewers for their valuable contribution.

\bibliography{custom}

\begin{thebibliography}{20}
\expandafter\ifx\csname natexlab\endcsname\relax\def\natexlab#1{#1}\fi

\bibitem[{Ajjour et~al.(2017)Ajjour, Chen, Kiesel, Wachsmuth, and
  Stein}]{ajjour2017unit}
Yamen Ajjour, Wei-Fan Chen, Johannes Kiesel, Henning Wachsmuth, and Benno
  Stein. 2017.
\newblock Unit segmentation of argumentative texts.
\newblock In \emph{Proceedings of the 4th Workshop on Argument Mining}, pages
  118--128.

\bibitem[{Devlin et~al.(2019)Devlin, Chang, Lee, and
  Toutanova}]{devlin-etal-2019-bert}
Jacob Devlin, Ming-Wei Chang, Kenton Lee, and Kristina Toutanova. 2019.
\newblock \href {https://doi.org/10.18653/v1/N19-1423} {{BERT}: Pre-training of
  deep bidirectional transformers for language understanding}.
\newblock In \emph{Proceedings of the 2019 Conference of the North {A}merican
  Chapter of the Association for Computational Linguistics: Human Language
  Technologies, Volume 1 (Long and Short Papers)}, pages 4171--4186,
  Minneapolis, Minnesota. Association for Computational Linguistics.

\bibitem[{Eger et~al.(2017)Eger, Daxenberger, and
  Gurevych}]{eger-etal-2017-neural}
Steffen Eger, Johannes Daxenberger, and Iryna Gurevych. 2017.
\newblock \href {https://doi.org/10.18653/v1/P17-1002} {Neural end-to-end
  learning for computational argumentation mining}.
\newblock In \emph{Proceedings of the 55th Annual Meeting of the Association
  for Computational Linguistics (Volume 1: Long Papers)}, pages 11--22,
  Vancouver, Canada. Association for Computational Linguistics.

\bibitem[{Gardner et~al.(2020)Gardner, Artzi, Basmov, Berant, Bogin, Chen,
  Dasigi, Dua, Elazar, Gottumukkala et~al.}]{gardner2020evaluating}
Matt Gardner, Yoav Artzi, Victoria Basmov, Jonathan Berant, Ben Bogin, Sihao
  Chen, Pradeep Dasigi, Dheeru Dua, Yanai Elazar, Ananth Gottumukkala, et~al.
  2020.
\newblock Evaluating models’ local decision boundaries via contrast sets.
\newblock In \emph{Findings of the Association for Computational Linguistics:
  EMNLP 2020}, pages 1307--1323.

\bibitem[{Goel et~al.(2021)Goel, Rajani, Vig, Taschdjian, Bansal, and
  R{\'e}}]{goel-etal-2021-robustness}
Karan Goel, Nazneen~Fatema Rajani, Jesse Vig, Zachary Taschdjian, Mohit Bansal,
  and Christopher R{\'e}. 2021.
\newblock \href {https://doi.org/10.18653/v1/2021.naacl-demos.6} {Robustness
  gym: Unifying the {NLP} evaluation landscape}.
\newblock In \emph{Proceedings of the 2021 Conference of the North American
  Chapter of the Association for Computational Linguistics: Human Language
  Technologies: Demonstrations}, pages 42--55, Online. Association for
  Computational Linguistics.

\bibitem[{Habernal et~al.(2014)Habernal, Eckle-Kohler, and
  Gurevych}]{habernal2014argumentation}
Ivan Habernal, Judith Eckle-Kohler, and Iryna Gurevych. 2014.
\newblock Argumentation mining on the web from information seeking perspective.
\newblock In \emph{ArgNLP}.

\bibitem[{Lauscher et~al.(2021)Lauscher, Wachsmuth, Gurevych, and
  Glava{\v{s}}}]{lauscher2021scientia}
Anne Lauscher, Henning Wachsmuth, Iryna Gurevych, and Goran Glava{\v{s}}. 2021.
\newblock Scientia potentia est--on the role of knowledge in computational
  argumentation.
\newblock \emph{arXiv preprint arXiv:2107.00281}.

\bibitem[{Liu et~al.(2019)Liu, Ott, Goyal, Du, Joshi, Chen, Levy, Lewis,
  Zettlemoyer, and Stoyanov}]{liu2019roberta}
Yinhan Liu, Myle Ott, Naman Goyal, Jingfei Du, Mandar Joshi, Danqi Chen, Omer
  Levy, Mike Lewis, Luke Zettlemoyer, and Veselin Stoyanov. 2019.
\newblock Roberta: A robustly optimized bert pretraining approach.
\newblock \emph{arXiv e-prints}, pages arXiv--1907.

\bibitem[{Mayer et~al.(2020{\natexlab{a}})Mayer, Cabrio, and
  Villata}]{mayer2020transformer}
Tobias Mayer, Elena Cabrio, and Serena Villata. 2020{\natexlab{a}}.
\newblock Transformer-based argument mining for healthcare applications.
\newblock In \emph{ECAI 2020}, pages 2108--2115. IOS Press.

\bibitem[{Mayer et~al.(2020{\natexlab{b}})Mayer, Marro, Cabrio, and
  Villata}]{mayer2020generating}
Tobias Mayer, Santiago Marro, Elena Cabrio, and Serena Villata.
  2020{\natexlab{b}}.
\newblock Generating adversarial examples for topic-dependent argument
  classification 1.
\newblock In \emph{Computational Models of Argument}, pages 33--44. IOS Press.

\bibitem[{Moore and Rayson(2018)}]{moore-rayson-2018-bringing}
Andrew Moore and Paul Rayson. 2018.
\newblock \href {https://aclanthology.org/C18-1097} {Bringing replication and
  reproduction together with generalisability in {NLP}: Three reproduction
  studies for target dependent sentiment analysis}.
\newblock In \emph{Proceedings of the 27th International Conference on
  Computational Linguistics}, pages 1132--1144, Santa Fe, New Mexico, USA.
  Association for Computational Linguistics.

\bibitem[{Niculae et~al.(2017)Niculae, Park, and
  Cardie}]{niculae-etal-2017-argument}
Vlad Niculae, Joonsuk Park, and Claire Cardie. 2017.
\newblock \href {https://doi.org/10.18653/v1/P17-1091} {Argument mining with
  structured {SVM}s and {RNN}s}.
\newblock In \emph{Proceedings of the 55th Annual Meeting of the Association
  for Computational Linguistics (Volume 1: Long Papers)}, pages 985--995,
  Vancouver, Canada. Association for Computational Linguistics.

\bibitem[{Niven and Kao(2019)}]{niven-kao-2019-probing}
Timothy Niven and Hung-Yu Kao. 2019.
\newblock \href {https://doi.org/10.18653/v1/P19-1459} {Probing neural network
  comprehension of natural language arguments}.
\newblock In \emph{Proceedings of the 57th Annual Meeting of the Association
  for Computational Linguistics}, pages 4658--4664, Florence, Italy.
  Association for Computational Linguistics.

\bibitem[{Poudyal et~al.(2020)Poudyal, Savelka, Ieven, Moens, Goncalves, and
  Quaresma}]{poudyal-etal-2020-echr}
Prakash Poudyal, Jaromir Savelka, Aagje Ieven, Marie~Francine Moens, Teresa
  Goncalves, and Paulo Quaresma. 2020.
\newblock \href {https://aclanthology.org/2020.argmining-1.8} {{ECHR}: Legal
  corpus for argument mining}.
\newblock In \emph{Proceedings of the 7th Workshop on Argument Mining}, pages
  67--75, Online. Association for Computational Linguistics.

\bibitem[{Reimers and Gurevych(2019)}]{reimers2019sentence}
Nils Reimers and Iryna Gurevych. 2019.
\newblock Sentence-bert: Sentence embeddings using siamese bert-networks.
\newblock In \emph{Proceedings of the 2019 Conference on Empirical Methods in
  Natural Language Processing and the 9th International Joint Conference on
  Natural Language Processing (EMNLP-IJCNLP)}, pages 3982--3992.

\bibitem[{Reuver et~al.(2021)Reuver, Verberne, Morante, and
  Fokkens}]{reuver-etal-2021-stance}
Myrthe Reuver, Suzan Verberne, Roser Morante, and Antske Fokkens. 2021.
\newblock \href {https://doi.org/10.18653/v1/2021.argmining-1.5} {Is stance
  detection topic-independent and cross-topic generalizable? - a reproduction
  study}.
\newblock In \emph{Proceedings of the 8th Workshop on Argument Mining}, pages
  46--56, Punta Cana, Dominican Republic. Association for Computational
  Linguistics.

\bibitem[{Ruiz-Dolz et~al.(2021)Ruiz-Dolz, Alemany, Barber{\'a}, and
  Garc{\'\i}a-Fornes}]{ruiz2021transformer}
Ramon Ruiz-Dolz, Jose Alemany, Stella M~Heras Barber{\'a}, and Ana
  Garc{\'\i}a-Fornes. 2021.
\newblock Transformer-based models for automatic identification of argument
  relations: A cross-domain evaluation.
\newblock \emph{IEEE Intelligent Systems}, 36(6):62--70.

\bibitem[{Schiller et~al.(2021)Schiller, Daxenberger, and
  Gurevych}]{schiller2021stance}
Benjamin Schiller, Johannes Daxenberger, and Iryna Gurevych. 2021.
\newblock Stance detection benchmark: How robust is your stance detection?
\newblock \emph{KI-K{\"u}nstliche Intelligenz}, 35(3):329--341.

\bibitem[{Trautmann et~al.(2020)Trautmann, Daxenberger, Stab, Sch{\"u}tze, and
  Gurevych}]{trautmann2020fine}
Dietrich Trautmann, Johannes Daxenberger, Christian Stab, Hinrich Sch{\"u}tze,
  and Iryna Gurevych. 2020.
\newblock Fine-grained argument unit recognition and classification.
\newblock In \emph{Proceedings of the AAAI Conference on Artificial
  Intelligence}, volume~34, pages 9048--9056.

\bibitem[{Zhang et~al.(2020)Zhang, Sheng, Alhazmi, and
  Li}]{zhang2020adversarial}
Wei~Emma Zhang, Quan~Z Sheng, Ahoud Alhazmi, and Chenliang Li. 2020.
\newblock Adversarial attacks on deep-learning models in natural language
  processing: A survey.
\newblock \emph{ACM Transactions on Intelligent Systems and Technology (TIST)},
  11(3):1--41.

\end{thebibliography}
\bibliographystyle{acl_natbib}

\appendix

\section{Appendix}\label{sec:appendix}
\paragraph{Software and Hyperparameters}
Our code and data are available at \url{github.com/jbkamp/repo-Rob-Token-AUR}. The implementation for the token-based model was retrieved from \url{github.com/trtm/AURC}, and we adapted it to train a sentence-based model through the \texttt{transformers. BertForSequenceClassification} class, at \url{huggingface.co}. For both, we used the large cased pre-trained model with whole word masking at \url{huggingface.co}. We used the same settings across models: learning rate was kept at 1e-5, dropout rate at 0.1 and the maximum length of the tokenised BERT input was set at 64 tokens. Optimizer adopted: AdamW. The batch size was set at 32 and models were trained for a maximum of 100 epochs with early stopping if the performance did not improve significantly after the 10th epoch.  \\

\paragraph{Dataset Versioning} \citet{trautmann2020fine} published their results based on the AURC-8 dataset, requested and obtained via e-mail correspondence. A second version at \url{github.com/trtm/AURC/tree/master/data} of the AURC-8 dataset was uploaded in a later moment with, \textit{cleaner parsing and encoding} (\url{github.com/trtm/AURC\#readme}, last consulted on June 16th 2022) but with the same number of labels and sentences. The two datasets differ to a low degree: $4.91\%$ of the sentences are not equal ($n=393$). Of this subset, all elements show a better cleaning of punctuation tokens compared to the original. To the best of our knowledge, this is the only difference between the original and the updated dataset. Therefore, we prefer using the updated, cleaner version of the dataset. We remove duplicate sentences within and across training set, development set and test set, per split (see resulting counts in Table \ref{fig:AURCstatistics}). \\

\paragraph{Segment-F1} In order to compute the segment-F1 score, we average over all sentence-wise segment-F1 scores, for each sentence in the evaluation set. To obtain a sentence-wise segment-F1 score we consider all pairs $<y,\hat{y}>$, where $y$ is the sequence of true labels for a segment and $\hat{y}$ is the sequence of predicted labels for that segment. Let $r$ be the overlap ratio between $y$ and $\hat{y}$:
\begin{equation}
    r = \frac{|y \cap \hat{y}|}{|y|}
\end{equation}
We only compute $r$ for segments where the label of $y$ is $PRO$ or $CON$. If $r > .5$ and labels are the same, $\hat{y}$ is considered a true prediction; otherwise, a false prediction. The sentence-wise segment-F1 is the number of true predictions over all predictions for that sentence. If the sentence does not contain $PRO$ nor $CON$ segments, and no $PRO$ nor $CON$ is predicted, the segment-F1 score for the sentence is 1.0. See Table \ref{fig:reproductionresultscomplete} for a full overview of the results, including token-F1, segment-F1 and sentence-F1.

\begin{table*}
\centering
\begin{tabular}{llrrrrrr}
\toprule
\multicolumn{2}{l}{} & \multicolumn{3}{c}{\textbf{\small in-domain}} & \multicolumn{3}{c}{\textbf{\small cross-domain}} \\ \midrule
\multicolumn{1}{l}{\textbf{\small \#}} & \textbf{\small topic} & \multicolumn{1}{r}{\textbf{\small train}} & \multicolumn{1}{r}{\textbf{\small dev}} & \textbf{\small test} & \multicolumn{1}{r}{\textbf{\small train}} & \multicolumn{1}{r}{\textbf{\small dev}} & \textbf{\small test} \\ \midrule
\multicolumn{1}{l}{\small 1} & {\small abortion} & \multicolumn{1}{r}{{\small 700}} & \multicolumn{1}{r}{{\small 99}} & {\small 200} & \multicolumn{1}{r}{{\small 800}} & \multicolumn{1}{r}{\small 0} & {\small 0} \\ 
\multicolumn{1}{l}{\small 2} & {\small cloning} & \multicolumn{1}{r}{{\small 696}} & \multicolumn{1}{r}{{\small 100}} & {\small 200} & \multicolumn{1}{r}{{\small 800}} & \multicolumn{1}{r}{\small 0} & {\small 0} \\ 
\multicolumn{1}{l}{\small 3} & {\small marijuana legalization} & \multicolumn{1}{r}{{\small 699}} & \multicolumn{1}{r}{{\small 100}} & {\small 200} & \multicolumn{1}{r}{{\small 800}} & \multicolumn{1}{r}{\small 0} & {\small 0} \\
\multicolumn{1}{l}{\small 4} & {\small minimum wage} & \multicolumn{1}{r}{{\small 699}} & \multicolumn{1}{r}{{\small 100}} & {\small 200} & \multicolumn{1}{r}{{\small 800}} & \multicolumn{1}{r}{\small 0} & {\small 0} \\ 
\multicolumn{1}{l}{\small 5} & {\small nuclear energy} & \multicolumn{1}{r}{{\small 699}} & \multicolumn{1}{r}{{\small 100}} & {\small 200} & \multicolumn{1}{r}{{\small 800}} & \multicolumn{1}{r}{\small 0} & {\small 0} \\ 
\multicolumn{1}{l}{\small 6} & {\small death penalty} & \multicolumn{1}{r}{{\small 700}} & \multicolumn{1}{r}{{\small 100}} & {\small 200} & \multicolumn{1}{r}{\small 0} & \multicolumn{1}{r}{{\small 800}} & {\small 0} \\ 
\multicolumn{1}{l}{\small 7} & {\small gun control} & \multicolumn{1}{r}{\small 0} & \multicolumn{1}{r}{\small 0} & {\small 0} & \multicolumn{1}{r}{\small 0} & \multicolumn{1}{r}{\small 0} & {\small 1,000} \\
\multicolumn{1}{l}{\small 8} & {\small school uniforms} & \multicolumn{1}{r}{\small 0} & \multicolumn{1}{r}{\small 0} & {\small 0} & \multicolumn{1}{r}{\small 0} & \multicolumn{1}{r}{\small 0} & {\small 1,000} \\ 
\bottomrule
\end{tabular}
\caption{The eight topics from the AURC-8 dataset \citep{trautmann2020fine} along with the number of sentence instances per data split after duplicates removal.}
\label{fig:AURCstatistics}
\end{table*}

\begin{table*}
\centering
\begin{tabular}{ccccccccccccc}
\toprule
\multicolumn{1}{l}{} & \multicolumn{4}{c}{\textbf{\small token-F1}} & \multicolumn{4}{c}{\textbf{\small segment-F1}} & \multicolumn{4}{c}{\textbf{\small sentence-F1}} \\ \cline{2-13} 
\multicolumn{1}{l}{\multirow{-2}{*}{\textbf{}}} & \multicolumn{2}{c}{\textit{\begin{tabular}[c]{@{}c@{}}{\small token}\\ {\small -based setup}\end{tabular}}} & \multicolumn{2}{c}{\textit{\begin{tabular}[c]{@{}c@{}}{\small sentence}\\ {\small -based setup}\end{tabular}}} & \multicolumn{2}{c}{\textit{\begin{tabular}[c]{@{}c@{}}{\small token}\\ {\small -based setup}\end{tabular}}} & \multicolumn{2}{c}{\textit{\begin{tabular}[c]{@{}c@{}}{\small sentence}\\ {\small -based setup}\end{tabular}}} & \multicolumn{2}{c}{\textit{\begin{tabular}[c]{@{}c@{}}{\small token}\\ {\small -based setup}\end{tabular}}} & \multicolumn{2}{c}{\textit{\begin{tabular}[c]{@{}c@{}}{\small sentence}\\ {\small -based setup}\end{tabular}}} \\ \midrule
\textit{\small model} & \multicolumn{1}{c}{\textbf{\small dev}} & \multicolumn{1}{c}{\textbf{\small test}} & \multicolumn{1}{c}{\textbf{\small dev}} & \textbf{\small test} & \multicolumn{1}{c}{\textbf{\small dev}} & \multicolumn{1}{c}{\textbf{\small test}} & \multicolumn{1}{c}{\textbf{\small dev}} & \textbf{\small test} & \multicolumn{1}{c}{\textbf{\small dev}} & \multicolumn{1}{c}{\textbf{\small test}} & \multicolumn{1}{c}{\textbf{\small dev}} & \textbf{\small test} \\ \midrule
\begin{tabular}[c]{@{}c@{}}{\small in-domain}\\ {\small BERT}{\tiny LARGE}\end{tabular} & \multicolumn{1}{c}{\small.732} & \multicolumn{1}{c}{\small.683} & \multicolumn{1}{c}{\small.671} & {\small.627} & \multicolumn{1}{c}{\small.749} & \multicolumn{1}{c}{\small.709} & \multicolumn{1}{c}{\small.599} & {\small.567} & \multicolumn{1}{c}{\small.738} & \multicolumn{1}{c}{\small.709} & \multicolumn{1}{c}{\small.759} & {\small.715} \\ \midrule
\begin{tabular}[c]{@{}c@{}}{\small in-domain}\\ {\small BERT}{\tiny LARGE}{\small+CRF}\end{tabular} & \multicolumn{1}{c}{\small.743} & \multicolumn{1}{c}{\small.696} & \multicolumn{1}{c}{\small.637} & {\small.622} & \multicolumn{1}{c}{\small.750} & \multicolumn{1}{c}{\small.724} & \multicolumn{1}{c}{\small.552} & {\small.547} & \multicolumn{1}{c}{\small.744} & \multicolumn{1}{c}{\small.711} & \multicolumn{1}{c}{\small.731} & {\small.725} \\ \midrule
\rowcolor[HTML]{FFFFC7} 
\begin{tabular}[c]{@{}c@{}}{\small in-domain}\\ {\small BERT}{\tiny LARGE}\end{tabular} & \multicolumn{1}{c}{\cellcolor[HTML]{FFFFC7}\begin{tabular}[c]{@{}c@{}}{\small.717}\\ {\tiny (.004)}\end{tabular}} & \multicolumn{1}{c}{\cellcolor[HTML]{FFFFC7}\begin{tabular}[c]{@{}c@{}}{\small.698}\\ {\tiny (.003)}\end{tabular}} & \multicolumn{1}{c}{\cellcolor[HTML]{FFFFC7}\begin{tabular}[c]{@{}c@{}}{\small.628}\\ {\tiny (.005)}\end{tabular}} & \begin{tabular}[c]{@{}c@{}}\textbf{\small.614}\\ {\tiny (.008)}\end{tabular} & \multicolumn{1}{c}{\cellcolor[HTML]{FFFFC7}\begin{tabular}[c]{@{}c@{}}{\small.776}\\ {\tiny (.011)}\end{tabular}} & \multicolumn{1}{c}{\cellcolor[HTML]{FFFFC7}\begin{tabular}[c]{@{}c@{}}{\small.749}\\ {\tiny (.005)}\end{tabular}} & \multicolumn{1}{c}{\cellcolor[HTML]{FFFFC7}\begin{tabular}[c]{@{}c@{}}{\small.514}\\ {\tiny (.009)}\end{tabular}} & \begin{tabular}[c]{@{}c@{}}{\small.500}\\ {\tiny (.004)}\end{tabular} & \multicolumn{1}{c}{\cellcolor[HTML]{FFFFC7}\begin{tabular}[c]{@{}c@{}}{\small.715}\\ {\tiny (.008)}\end{tabular}} & \multicolumn{1}{c}{\cellcolor[HTML]{FFFFC7}\begin{tabular}[c]{@{}c@{}}\textbf{\small.708}\\ {\tiny (.004)}\end{tabular}} & \multicolumn{1}{c}{\cellcolor[HTML]{FFFFC7}\begin{tabular}[c]{@{}c@{}}{\small.726}\\ {\tiny (.007)}\end{tabular}} & \begin{tabular}[c]{@{}c@{}}\textbf{\small.713}\\ {\tiny (.012)}\end{tabular} \\ \midrule
\rowcolor[HTML]{FFFFC7} 
\begin{tabular}[c]{@{}c@{}}{\small in-domain}\\ {\small BERT}{\tiny LARGE}{\small+CRF}\end{tabular} & \multicolumn{1}{c}{\cellcolor[HTML]{FFFFC7}\begin{tabular}[c]{@{}c@{}}{\small.716}\\ {\tiny (.003)}\end{tabular}} & \multicolumn{1}{c}{\cellcolor[HTML]{FFFFC7}\begin{tabular}[c]{@{}c@{}}\textbf{\small.{\small 696}}\\ {\tiny (.003)}\end{tabular}} & \multicolumn{1}{c}{\cellcolor[HTML]{FFFFC7}-} & - & \multicolumn{1}{c}{\cellcolor[HTML]{FFFFC7}\begin{tabular}[c]{@{}c@{}}{\small.766}\\ {\tiny (.003)}\end{tabular}} & \multicolumn{1}{c}{\cellcolor[HTML]{FFFFC7}\begin{tabular}[c]{@{}c@{}}{\small.743}\\ {\tiny (.008)}\end{tabular}} & \multicolumn{1}{c}{\cellcolor[HTML]{FFFFC7}-} & - & \multicolumn{1}{c}{\cellcolor[HTML]{FFFFC7}\begin{tabular}[c]{@{}c@{}}{\small.718}\\ {\tiny (.008)}\end{tabular}} & \multicolumn{1}{c}{\cellcolor[HTML]{FFFFC7}\begin{tabular}[c]{@{}c@{}}\textbf{\small.711}\\ {\tiny (.006)}\end{tabular}} & \multicolumn{1}{c}{\cellcolor[HTML]{FFFFC7}-} & - \\ \midrule
\begin{tabular}[c]{@{}c@{}}{\small cross-domain}\\ {\small BERT}{\tiny LARGE}\end{tabular} & \multicolumn{1}{c}{\small.604} & \multicolumn{1}{c}{\small.596} & \multicolumn{1}{c}{\small.550} & {\small.544} & \multicolumn{1}{c}{\small.653} & \multicolumn{1}{c}{\small.626} & \multicolumn{1}{c}{\small.487} & {\small.473} & \multicolumn{1}{c}{\small.606} & \multicolumn{1}{c}{\small.598} & \multicolumn{1}{c}{\small.628} & {\small.602} \\ \midrule
\begin{tabular}[c]{@{}c@{}}{\small cross-domain}\\ {\small BERT}{\tiny LARGE}{\small+CRF}\end{tabular} & \multicolumn{1}{c}{\small.615} & \multicolumn{1}{c}{\small.620} & \multicolumn{1}{c}{\small.505} & {\small.519} & \multicolumn{1}{c}{\small.681} & \multicolumn{1}{c}{\small.649} & \multicolumn{1}{c}{\small.456} & {\small.464} & \multicolumn{1}{c}{\small.627} & \multicolumn{1}{c}{\small.610} & \multicolumn{1}{c}{\small.569} & {\small.573} \\ \midrule
\rowcolor[HTML]{FFFFC7} 
\begin{tabular}[c]{@{}c@{}}{\small cross-domain}\\ {\small BERT}{\tiny LARGE}\end{tabular} & \multicolumn{1}{c}{\cellcolor[HTML]{FFFFC7}\begin{tabular}[c]{@{}c@{}}{\small.581}\\ {\tiny (.011)}\end{tabular}} & \multicolumn{1}{c}{\cellcolor[HTML]{FFFFC7}\begin{tabular}[c]{@{}c@{}}\textbf{\small.587}\\ {\tiny (.008)}\end{tabular}} & \multicolumn{1}{c}{\cellcolor[HTML]{FFFFC7}\begin{tabular}[c]{@{}c@{}}{\small.515}\\ {\tiny (.012)}\end{tabular}} & \begin{tabular}[c]{@{}c@{}}\textbf{\small.529}\\ {\tiny (.011)}\end{tabular} & \multicolumn{1}{c}{\cellcolor[HTML]{FFFFC7}\begin{tabular}[c]{@{}c@{}}{\small.630}\\ {\tiny (.007)}\end{tabular}} & \multicolumn{1}{c}{\cellcolor[HTML]{FFFFC7}\begin{tabular}[c]{@{}c@{}}{\small.603}\\ {\tiny (.011)}\end{tabular}} & \multicolumn{1}{c}{\cellcolor[HTML]{FFFFC7}\begin{tabular}[c]{@{}c@{}}{\small.424}\\ {\tiny (.014)}\end{tabular}} & \begin{tabular}[c]{@{}c@{}}{\small.433}\\ {\tiny (.004)}\end{tabular} & \multicolumn{1}{c}{\cellcolor[HTML]{FFFFC7}\begin{tabular}[c]{@{}c@{}}\textbf{\small.591}\\ {\tiny (.016)}\end{tabular}} & \multicolumn{1}{c}{\cellcolor[HTML]{FFFFC7}\begin{tabular}[c]{@{}c@{}}\textbf{\small.604}\\ {\tiny (.009)}\end{tabular}} & \multicolumn{1}{c}{\cellcolor[HTML]{FFFFC7}\begin{tabular}[c]{@{}c@{}}{\small.596}\\ {\tiny (.010)}\end{tabular}} & \begin{tabular}[c]{@{}c@{}}{\small.566}\\ {\tiny (.017)}\end{tabular} \\ \midrule
\rowcolor[HTML]{FFFFC7} 
\begin{tabular}[c]{@{}c@{}}{\small cross-domain}\\ {\small BERT}{\tiny LARGE}{\small+CRF}\end{tabular} & \multicolumn{1}{c}{\cellcolor[HTML]{FFFFC7}\begin{tabular}[c]{@{}c@{}}{\small.584}\\ {\tiny (.009)}\end{tabular}} & \multicolumn{1}{c}{\cellcolor[HTML]{FFFFC7}\begin{tabular}[c]{@{}c@{}}{\small.578}\\ {\tiny (.008)}\end{tabular}} & \multicolumn{1}{c}{\cellcolor[HTML]{FFFFC7}-} & - & \multicolumn{1}{c}{\cellcolor[HTML]{FFFFC7}\begin{tabular}[c]{@{}c@{}}{\small.627}\\ {\tiny (.012)}\end{tabular}} & \multicolumn{1}{c}{\cellcolor[HTML]{FFFFC7}\begin{tabular}[c]{@{}c@{}}{\small.593}\\ {\tiny (.004)}\end{tabular}} & \multicolumn{1}{c}{\cellcolor[HTML]{FFFFC7}-} & - &
\multicolumn{1}{c}{\cellcolor[HTML]{FFFFC7}\begin{tabular}[c]{@{}c@{}}{\small.601}\\ {\tiny (.011)}\end{tabular}} & \multicolumn{1}{c}{\cellcolor[HTML]{FFFFC7}\begin{tabular}[c]{@{}c@{}}\textbf{\small.609}\\ {\tiny (.007)}\end{tabular}} & \multicolumn{1}{c}{\cellcolor[HTML]{FFFFC7}-} & - \\ \bottomrule
\end{tabular}
\caption{Full overview of the original results (white background) compared to reproduction results (non-white background). Models are divided over an in-domain setting and a cross-domain setting. Reproduction results show the mean scores from 5 runs, along with the standard deviation (within parentheses). The reproduction scores where the original score falls within two standard deviations from the mean are given in bold.}
\label{fig:reproductionresultscomplete}
\end{table*}
\end{document}